\documentclass[11pt]{article}

\usepackage{acl}

\usepackage{times}
\usepackage{latexsym}

\usepackage[T1]{fontenc}

\usepackage[utf8]{inputenc}

\usepackage{microtype}

\usepackage{inconsolata}

\usepackage{graphicx}
\usepackage{hyperref}
\usepackage{booktabs}
\usepackage{amsmath}
\usepackage{multirow}
\usepackage{xcolor}
\usepackage{algorithm}
\usepackage{algpseudocode}
\usepackage{alphabeta}
\usepackage{pifont}
\usepackage{amssymb}
\usepackage{tikz}
\usetikzlibrary{shapes.geometric, arrows, positioning, calc}

\newcommand{\xmark}{\ding{55}}
\newcommand{\warning}{\ding{43}}

\title{LLMs Got Rhythm? Hybrid Phonological Filtering for Greek Poetry Rhyme Detection and Generation}

\author{Stergios Chatzikyriakidis \\
  Department of Philology \\
  University of Crete \\
  \texttt{stergios.chatzikyriakidis@uoc.gr} \\\And
  Anastasia Natsina \\
  Department of Philology \\
  University of Crete \\
  \texttt{natsina@uoc.gr} \\}
\begin{document}
\maketitle

\begin{abstract}
Large Language Models (LLMs), despite their remarkable capabilities across NLP tasks, struggle with phonologically-grounded phenomena like rhyme detection and generation. This is even more evident in lower-resource languages such as Modern Greek. In this paper, we present a hybrid system that combines LLMs with deterministic phonological algorithms to achieve accurate rhyme identification/analysis and generation. Our approach implements a comprehensive taxonomy of Greek rhyme types, including Pure, Rich, Imperfect, Mosaic, and Identical Pre-rhyme Vowel (IDV) patterns, and employs an agentic generation pipeline with phonological verification. We evaluate multiple prompting strategies (zero-shot, few-shot, Chain-of-Thought, and RAG-augmented) across several LLMs including Claude 3.7 and 4.5, GPT-4o, Gemini 2.0 and open-weight models like Llama 3.1 8B and 70B and Mistral Large. Results reveal a significant "Reasoning Gap": while native-like models (Claude 3.7) perform intuitively (40\% accuracy in identification), reasoning-heavy models (Claude 4.5) achieve state-of-the-art performance (54\%) only when prompted with Chain-of-Thought. Most critically, pure LLM generation fails catastrophically (under 4\% valid poems), while our hybrid verification loop restores performance to 73.1\%. We release our system and a  corpus of 40,000+ rhymes, derived from the \textit{Anemoskala} and \textit{Interwar Poetry} corpora, to support future research.\footnote{The material needed to run the experiments and verify the results of this paper can be found here: \url{https://osf.io/d7hr6/overview?view_only=6e780f09bfd444ccb60ec3732b0d2a9b}}
\end{abstract}

\section{Introduction}

Rhyme is a fundamental feature of verse across cultures. In Modern Greek, rhyme (ομοιοκαταληξία) has been systematically employed from the Cretan Renaissance (14th--17th centuries) all the way to contemporary poetry and popular music, including hip-hop and rap \cite{topintzi2019quantifying}.

LLMs, unsurprisingly given their prevalent text-based training, exhibit notable weaknesses in phonologically-aware reasoning. LLMs process text at the token level, which poorly aligns with phonological units like syllables, stress patterns, and rhyme domains. 

We tackle the issue of rhyme identification and generation using a hybrid Neural-Symbolic architecture that combines the generative and reasoning capabilities of LLMs with deterministic phonological algorithms. Our contributions are:

\begin{enumerate}
    \item A dataset of 40k rhymes aggregated from existing Modern Greek corpora, based on the phonological taxonomy found in \cite{topintzi2019quantifying}.
    
    \item A hybrid detection system combining LLM prompting strategies with rule-based phonological verification, achieving 100\% verification accuracy compared to 54\% (best case) for pure LLM approaches.
    
    \item An agentic generation pipeline with a Generate-Verify-Refine loop that raises the validity of generated poems from nearly 0\% to 73.1\%.
    
    \item A multi-model evaluation across proprietary (Claude, GPT-4o) and open-weight models, with analysis of prompting strategies including RAG augmentation.
\end{enumerate}

\section{Background}

\subsection{Rhyme in Modern Greek Poetry}

Rhyme in Modern Greek poetry dates to the medieval period, appearing in works like those of Stefanos Sachlikis (14th c.), and flourishing in Cretan Renaissance masterpieces such as Kornaros's \textit{Erotokritos}. The phenomenon spans the Heptanesian Romanticism of Solomos, the Athenian school, Parnassian and Symbolist movements, and continues, albeit more selectively, in modern and contemporary poetry \cite{kokkolis1993}.

Greek rhyme exhibits several distinctive characteristics:

\paragraph{Stress-based Classification} Greek is a stress-accent language where rhyme domains are defined relative to the stressed syllable. More particularly, Greek is subject to the three-syllable rule according to which accent must fall in one of the last three syllables of the word (broadly defined, includes also cases of what we call phonological word, e.g. a word plus a weak object pronoun). We follow \citet{topintzi2019quantifying} and use the following categories:

\begin{itemize}
    \item \textbf{Masculine (M)}: Here, stress is on the final syllable (oxytone), e.g., καρδιά/φωτιά
    \item \textbf{Feminine-2 (F2)}: In this category, stress is on the penultimate syllable (paroxytone), e.g., θάλασσα/τάλασσα
    \item \textbf{Feminine-3 (F3)}: Here,  stress is on the antepenultimate syllable (proparoxytone), e.g., στόματα/σώματα
\end{itemize}

 Greek poetry further employs several other rhyme types that are independent of stress position:

\textbf{Rich Rhyme (RICH)}: In this type, the onset consonant(s) of the stressed syllable must match. This is further distinguished into Total Rich (TR) rhyme where complete onset matching is at play, or Partial Rich (PR) with partial matching:
\begin{itemize}
    \item TR-S (singleton): καλά/μαλά [ka-'la]/[ma-'la]
    \item PR-C1 (first consonant): στόματα/σώματα ['sto-ma-ta]/['so-ma-ta]
\end{itemize}

\textbf{Identical Pre-rhyme Vowel (IDV)}: In this category, the vowel that precedes the stressed syllable must match:
\begin{itemize}
    \item ξανθή/γραφή: pre-stress vowel [a] matches
\end{itemize}

\textbf{Mosaic (MOS)}: In this category, the rhyme domain spans across more than one word:
\begin{itemize}
    \item όνομά της/ο μπάτης: ['o-no-'ma tis]/[o 'ba-tis]
\end{itemize}

\textbf{Imperfect (IMP)}: In this category, [artial] phonetic matching with systematic variation:
\begin{itemize}
    \item IMP-V: Vowel differs (χάνετε/γίνετε)
    \item IMP-C: Consonant differs (ξαφνίζει/τεχνίτη)
    \item IMP-0F: Final consonant-zero alternation
\end{itemize}

\subsection{Computational Approaches to Rhyme}
Early approaches on rhyme identification used classic unsupervised machine learning  \citep{reddy2011unsupervised} or probabilistic models using phoneme frequencies for rhyme detection in rap music \citep{hirjee2010automatic}. More recent supervised methods based on neural networks achieve higher accuracies. For example,  \citet{haider2018supervised} achieve 97\% accuracy via a single Siamese Recurrent Network model trained in German, English, and French, using no explicit phonetic features. 

In poetry generation, we find early statistical machine translation for Classical Chinese quatrain generation by \citet{he2012generating}, where they in effect treat each line as a kind of translation of the previous line. The Hafez system  \citep{ghazvininejad2016generating} is a hybrid system that puts together  finite-state acceptors that encode metrical and rhyme constraints with RNNs for English sonnet generation. \citet{lau2018deep} developed Deep-speare, a joint neural model for Shakespearean sonnets whose outputs proved largely indistinguishable from human verse in crowd evaluations. Moving on to the LLM era, we find byte-level transformers such as ByGPT5 \citep{belouadi2023bygpt5}, as well as synthetic-data approaches like GPoeT \citep{popescu2023gpoet}.

For Greek specifically, the only computational work is the one by \citet{topintzi2019quantifying}, which resulted in the Greek Rhyme database (GrRh). The authors use  rule-based algorithms that cover multiple rhyme types, though with acknowledged limitations in precision. This is a pioneering paper that combines solid theoretical linguistics knowledge with computational implementation and its theoretical core is one of the inspirations of this paper.

\section{Dataset}
We constructed our dataset by aggregating and standardizing two primary high-quality digital resources for Modern Greek poetry. The first source is the Anemoskala archive from the Centre for the Greek Language (KEG - Κέντρο Ελληνικής Γλώσσας), which provides extensive digitized collections of major poets. The second source is the Interwar Poetry Dataset, an open-access dataset created by Dr Natsina and Professor Chatzikyriakidis with the help of undergraduate students at the Philology Department, University of Crete\footnote{Dataset available at: \url{https://github.com/StergiosChatzikyriakidis/Modern_Greek_Literature/tree/v1}.} This corpus comprises over 600 poems by interwar Greek poets. We merged these collections, normalized the JSON format, and applied our phonological cleaning pipeline to ensure consistency. The corpus statistics are shown in \ref{tab:corpus_clean}.

\begin{table}[t]
\centering
\small
\begin{tabular}{lr}
\toprule
\textbf{Poet} & \textbf{Rhyme Pairs} \\
\midrule
Palamas & 20,620 \\
Tellos Agras & 6,119 \\
Valaoritis & 4,202 \\
Solomos & 2,518 \\
Karyotakis & 1,692 \\
Cavafy & 1,585 \\
Fotos Giofyllis & 1,565 \\
Kostas Ouranis & 792 \\
Napoleon Lapathiotis & 495 \\
Romos Filiras & 484 \\
Mitsos Papanikolaou & 404 \\
Kalvos & 100 \\
\midrule
\textbf{Total} & \textbf{40,576} \\
\bottomrule
\end{tabular}
\caption{Corpus composition by poet.}
\label{tab:corpus_composition}
\end{table}

\section{System Architecture}

\begin{figure*}[t]
\centering
\begin{tikzpicture}[
    node distance=0.8cm and 1.5cm,
    auto,
    font=\small\sffamily,
    block/.style={rectangle, draw, fill=blue!10, text width=7em, text centered, rounded corners, minimum height=3em},
    llm/.style={rectangle, draw, fill=orange!10, text width=7em, text centered, rounded corners, minimum height=3em},
    decision/.style={diamond, draw, fill=green!10, text width=4em, text centered, inner sep=0pt, aspect=2},
    line/.style={draw, -latex, thick},
    cloud/.style={draw, ellipse, fill=red!10, minimum height=2em, text width=4em, text centered},
    label/.style={font=\bfseries\large}
]

  \node [block, fill=blue!20] (engine) {\textbf{Phonological Engine}};
  \node [block, below=0.8cm of engine, text width=8em, fill=gray!10, dashed] (rules) {Symbolic Rules\\(Syllabification/Stress)};

  \node [cloud, above left=1.2cm and 2.5cm of engine] (input_text) {Input Pair};
  \node [llm, below=0.6cm of input_text] (llm_id) {LLM Analysis};
  \node [decision, below=0.6cm of llm_id] (compare) {Correct?};
  \node [block, below=0.8cm of compare, fill=purple!10, text width=6em] (result) {Validated Label};
  
  \node [label, above=0.3cm of input_text] {A. Identification};

  \node [cloud, above right=1.2cm and 2.5cm of engine] (prompt) {Prompt};
  \node [llm, below=0.6cm of prompt] (llm_gen) {LLM};
  \node [decision, below=0.6cm of llm_gen] (check) {Valid?};
  \node [cloud, below=1.0cm of check] (output) {Poem};
  \node [block, right=0.8cm of check, text width=5em, font=\scriptsize] (feedback) {Feedback Gen};
  
  \node [label, above=0.3cm of prompt] {B. Generation};

  \path [line, dashed] (rules) -- (engine);

  \path [line] (input_text) -- (llm_id);
  \path [line] (llm_id) -- (compare);
  \path [line, blue] (engine) -- node[midway, above, font=\scriptsize, sloped] {Ground Truth} (compare);
  \path [line] (input_text) -- node[midway, right, font=\scriptsize] {} (engine); 
  \path [line] (compare) -- (result);

  \path [line] (prompt) -- (llm_gen);
  \path [line] (llm_gen) -- node[right, font=\scriptsize] {Draft} (check);
  
  \path [line, blue] (engine) -- node[midway, above, font=\scriptsize, sloped] {Verify} (check);

  \path [line] (check) -- node[right] {Yes} (output);
  \path [line] (check) -- node[above] {No} (feedback);
  \path [line] (feedback) |- (llm_gen);

  \draw[dashed, gray] ($(engine.north)+(0, 2.0)$) -- ($(engine.south)+(0, -3.0)$);

\end{tikzpicture}
\caption{Hybrid system architecture. \textbf{Left}: Identification combines LLM predictions with Engine-generated ground truth for validation. \textbf{Right}: Generation uses the Engine to verify and refine LLM outputs.}
\label{fig:architecture}
\end{figure*}
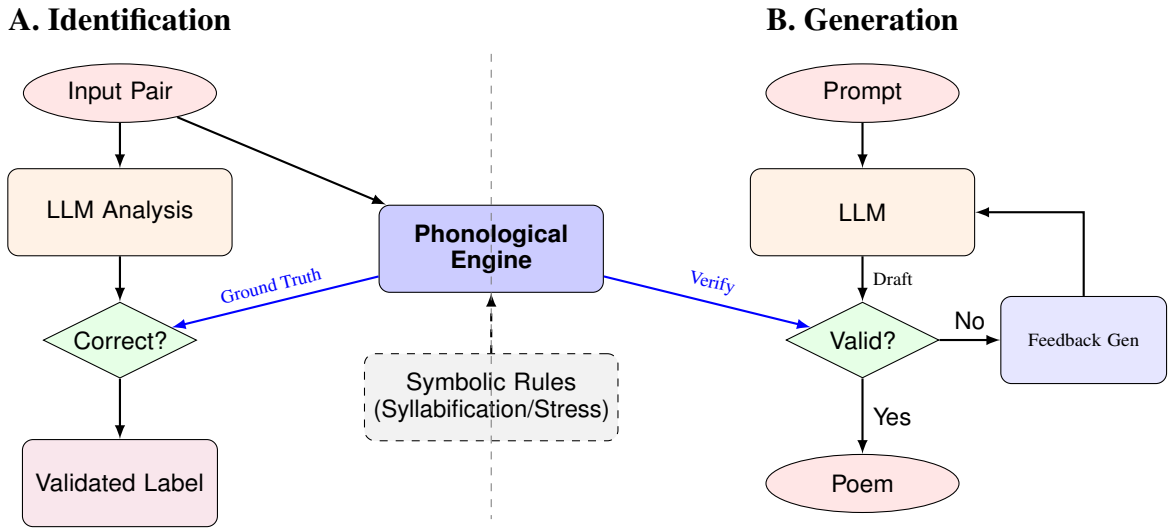

We implement a hybrid, neural-symbolic architecture thatr combines LLM capabilities with deterministic symbolic phonological rules (Figure~\ref{fig:architecture})

\subsection{Phonological Engine}
The basis of the phonological symbolic processor/verifier  is based on \cite{topintzi2019quantifying}. Its purpose is exactly  to handle Greek-specific rhyme analysis:

\begin{itemize}
    \item \textbf{Syllabification}.A number of Greek syllabification rules are implemented. The algorithm is designed to identify syllable boundaries based on Greek-specific phonotactic constraints.
    \item \textbf{Stress Detection}. We use the accent marks found in Greek orthography (ά, έ, ή, ί, ό, ύ, ώ) in orcer to identify stress. In case of words with clitics (notably weak object pronouns, e.g., κάλεσέ με), we have a mechanism to handle stress domain extension.
    \item \textbf{Rhyme Domain Extraction}. The domain  of the rhyme extends from the stressed vowel until the end of the phonological phrase. In the case of Mosaic rhymes, it can span multiple orthographic words.
    \item  \textbf{Phonetic Transcription}. We convert Greek orthography to a phonetic representation.

\end{itemize}

\subsection{Rhyme Classification Module}

We develop a classification module based on the phonological module. It compares rhyme domain pairs and assigns labels. More specifically it checks:

\begin{enumerate}
    \item \textbf{Position Classification}: Here it determines where the stress falls and classifies as M/F2/F3.
    \item \textbf{Perfect Match Check}: In this part, the post-stress material is checked to see if it is an exact  match.
    \item \textbf{Feature Detection}: This checks  for RICH rhyme (onset matching), IDV rhyme (pre-stress vowel) and/or Mosaic (MOS) rhyme (word boundary crossing).
    \item \textbf{Imperfection Analysis}: In case the rhyme is not perfect, classify accordingly (IMP-V, IMP-C, IMP-0F, IMP-0M).
\end{enumerate}

The output result is a compound label such as \texttt{F2-TR-S-IDV} (Feminine-2, Total Rich Singleton, Identical pre-rhyme Vowel).

\subsection{LLM Integration}
We integrate a number of LLM providers through a unified API layer, including Anthropic (Claude Sonnet 3.7/4.5), OpenAI (GPT-4o), Gemini and Open Models (Llama 3.1 8b and 70b and Mistral Large). We use both open and closed models and models of varying sizes and reported capabilities. 

 We implement five prompting strategies for rhyme identification:
\begin{itemize}
    \item \textbf{Zero-Shot Structured}: Provides the rhyme taxonomy and requests analysis
    \item \textbf{Zero-Shot CoT}: Requests explicit reasoning through detection steps
    \item \textbf{RAG Augmentation}: Retrieves relevant examples from our corpus
\end{itemize}

\subsection{Agentic Generation Pipeline}

For the task of rhyme generation, we implement an agentic loop with phonological verification based on our phonological verifier:

\begin{algorithm}[t]
\caption{Generate-Verify-Refine Loop}
\begin{algorithmic}[1]
\small
\State \textbf{Input:} Theme, rhyme\_type, features, num\_lines
\State \textbf{Output:} Phonologically valid poem
\State $attempts \gets 0$
\While{$attempts < 15$}
    \State $poem \gets LLM.generate(prompt)$
    \State $errors \gets verify\_rhymes(poem)$
    \If{$errors = \emptyset$}
        \State \Return $poem$
    \Else
        \State $feedback \gets format\_errors(errors)$
        \State $prompt \gets update\_prompt(feedback)$
        \State $attempts \gets attempts + 1$
    \EndIf
\EndWhile
\State \Return $poem$ with warning
\end{algorithmic}
\end{algorithm}

The verifier's task is to  check each rhyme pair using the phonological rules and provide  feedback in case the rhyme does not comply with the rules.

\section{Experimental Setup}
 
 We use a test set of 40  poems for rhyme identification. This produces  a total of  160 test cases in total (2 strategies × 2 RAG configs × 40 poems). The set has a  balanced distribution across rhyme types (13 Masculine, 16 Feminine-2, 11 Feminine-3), as well as comprehensive coverage of rhyme features including more rare types of ryme (21 PURE, 10 RICH, 10 IDV, 6 IMPERFECT, 5 MOSAIC). We went for a  balance that was big enough  to make meaningful claims, while, at the same time, maintaining evaluation feasibility across 8 models and 4 configurations (1,280 total API calls). 

 For generation, we evaluate generation quality on 26 test cases with specified rhyme constraints (e.g., ``Write a 4-line poem with F3-RICH rhyme on theme: love''). Each test runs twice: once with our verification loop (Generate-Verify-Refine) and once without (pure LLM generation).

\section{Results}
 
\subsection{Rhyme Identification Results}

 In Table~\ref{tab:identification_all}, we see the  results across all prompting configurations.
 
 \begin{table*}[h!]
 \centering
 \small
 \begin{tabular}{lcccc}
 \toprule
 \textbf{Model} & \textbf{Structured} & \textbf{Structured+RAG} & \textbf{CoT} & \textbf{CoT+RAG} \\
 \midrule
 \multicolumn{5}{l}{\textit{Proprietary Models}} \\
 Claude 4.5 & 53.8\% & 26.9\% & 46.2\% & \textbf{53.8\%} \\
 Claude 3.7 & 38.5\% & \textbf{42.3\%} & 42.3\% & 30.8\% \\
 GPT-4o & 7.7\% & 7.7\% & \textbf{50.0\%} & 26.9\% \\
 Gemini 2.0 & 23.1\% & \textbf{42.3\%} & 19.2\% & 11.5\% \\
 \midrule
 \multicolumn{5}{l}{\textit{Open-Weight Models}} \\
 Mistral Large & 11.5\% & 15.4\% & \textbf{26.9\%} & 11.5\% \\
 Llama 3.1 70B & \textbf{23.1\%} & 19.2\% & 11.5\% & 15.4\% \\
 Llama 3.3 70B & \textbf{23.1\%} & 23.1\% & 7.7\% & 7.7\% \\
 Llama 3.1 8B & 7.7\% & 15.4\% & 7.7\% & 3.8\% \\
 \bottomrule
 \end{tabular}
 \caption{Rhyme identification accuracy (\%) across   configurations. Bold means best configuration per model.}
 \label{tab:identification_all}
 \end{table*}
 
 Table~\ref{tab:identification_by_type} breaks down performance by rhyme type, revealing systematic biases.
 
 \begin{table}[t]
 \centering
 \small
 \begin{tabular}{lccc}
 \toprule
 \textbf{Model} & \textbf{M (n=32)} & \textbf{F2 (n=40)} & \textbf{F3 (n=32)} \\
 \midrule
 Claude 4.5 & \textbf{65.6\%} & 47.5\% & 21.9\% \\
 Claude 3.7 & 37.5\% & \textbf{52.5\%} & 21.9\% \\
 GPT-4o & 31.2\% & 25.0\% & 12.5\% \\
 Gemini 2.0 & 43.8\% & 20.0\% & 9.4\% \\
 \midrule
 Mistral Large & 25.0\% & 17.5\% & 6.2\% \\
 Llama 3.1 70B & 12.5\% & 32.5\% & 3.1\% \\
 Llama 3.3 70B & 25.0\% & 12.5\% & 9.4\% \\
 Llama 3.1 8B & 6.2\% & 12.5\% & 6.2\% \\
 \bottomrule
 \end{tabular}
 \caption{Identification accuracy by rhyme type. All models struggle most with F3 (proparoxytone) rhymes.}
 \label{tab:identification_by_type}
 \end{table}
 
 Table~\ref{tab:feature_detection} shows feature detection accuracy for each individual feature type.
 
 \begin{table*}[t]
 \centering
 \small
 \begin{tabular}{lccccc}
 \toprule
 \textbf{Model} & \textbf{PURE (n=68)} & \textbf{RICH (n=24)} & \textbf{MOSAIC (n=4)} & \textbf{IDV (n=24)} & \textbf{IMP (n=8)} \\
 \midrule
 Mistral Large & 4.4\% & \textbf{20.8\%} & 0.0\% & 12.5\% & 12.5\% \\
 Llama 3.1 70B & 5.9\% & 16.7\% & 0.0\% & 12.5\% & \textbf{37.5\%} \\
 Claude 3.7 & 8.8\% & 12.5\% & 0.0\% & 0.0\% & 25.0\% \\
 Claude 4.5 & 5.9\% & 8.3\% & 0.0\% & 4.2\% & 25.0\% \\
 Llama 3.3 70B & 2.9\% & 8.3\% & 0.0\% & 0.0\% & \textbf{37.5\%} \\
 Gemini 2.0 & 5.9\% & 0.0\% & 0.0\% & 0.0\% & \textbf{37.5\%} \\
 GPT-4o & 5.9\% & 0.0\% & 0.0\% & 0.0\% & 25.0\% \\
 \bottomrule
 \end{tabular}
 \caption{Feature detection accuracy by individual feature type. \textbf{MOSAIC rhymes are never detected} by any model. RICH and IDV detection remains below 21\%.}
 \label{tab:feature_detection}
 \end{table*}
 
 A number of interesting findings are borne out from our experiments. First, Claude 4.5 exhibits large variation in performance (26.9\% to 53.8\%) between non-CoT and CoT modes. This might indicate that reasoning-heavy models may require explicit prompting strategies.
 
 Second,  we find that the best  performance from proprietary models (Claude 4.5 using CoT RAG: 53.8\%) is better than the best open model performance (Mistral Large using Chain-of-Thought: 26.9\%) by roughly 27 points. This is a substantial difference that points to a large difference in capability for low-resource phonological tasks.
     
 Furthermore, we see that the size  of the model plays a role. Llama 3.1 70B is significantly better from Llama 3.1 8B (average accuracy 16.2\% vs 8.8\% across configurations), demonstrating that the rhyme detection task in Modern Greek benefits from larger models.
     
 In terms of diffuclty with respect to stress, F3 rhymes seem to be the hardest, as all models achieve lowest accuracy on F3 (proparoxytone) rhymes, with most below 22\%. In terms opf prompting strategies, Chain-of-Thought is helps   GPT-4o substantiatlly, increasing its accuracy from 7.1\% (Structured) to 50.0\% (CoT). The same is true for Claude 4.5 achieves peak overall accuracy (53.8\%) with CoT RAG prompting. Open-weight models generally fail to leverage CoT effectively.
     
 Mosaic rhymes seem to be the most difficult to detect. No model achieved an exact feature match for MOSAIC rhymes (0\% accuracy). However, Claude 4.5 using CoT+RAG successfully identified the MOSAIC feature in qualitative analysis, marking it as the only model capable of this complex phonological parsing.
 
 \subsection{Qualitative Analysis}
 To better the model's capabilities and failure modes, we analyzed 9 representative test cases, checking the performance for all models and configurations (full  outputs are provided in Appendix A):
 
 \begin{enumerate}
     \item \textbf{Baseline Success (M-PURE)}:
     \textit{απαιτώ} / \textit{λαμπρό}.
     This was correctly identified exactly by Claude 4.5 (Structured+RAG). GPT-4o (CoT) got the rhyme type right, but missed the exact features. Llama 3.1 8B (all configurations) failed this case across the board.
 
     \item \textbf{Baseline Success (F2-PURE)}:
     \textit{κρίνοι} / \textit{κρίνει}.
     This ia a perfect homophone rhyme. It was correctly identified  by Llama 3.1 70B (Structured+RAG). Claude 4.5 (Structured) flagged this as 'COPY', conflating distinct-lemma homophony, with identical rhyme (repetition).
 
     \item \textbf{Structural Failure (F3)}:
     \textit{παράπονο} / \textit{άπονο}.
     No model achieved a perfect match. Claude 3.7 (Structured+RAG) and GPT-4o (CoT+RAG) correctly identified the rhyme type (F3).  Claude 4.5 failed to identify the rhyme type in all configurations.
 
     \item \textbf{Mosaic Failure}:
     \textit{λυγμέ} / \textit{για με}.
     Cross-boundary rhyme. Claude 3.7 (Structured+RAG) was the only configuration to get a perfect match. Claude 4.5 (CoT+RAG) successfully identified the MOSAIC feature, but added  extraneous tags. Open models consistently failed.
 
     \item \textbf{Feature Hallucination (Rich)}:
     \textit{αφρός} / \textit{εμπρός}.
     Correctly identified by Claude 4.5 (CoT+RAG) and Gemini 2.0 (Structured+RAG). Claude 4.5 (Structured+RAG) hallucinated a RICH tag in zero-shot. However, the CoT configuration corrected it by analyzing the phonetics (/fr/ vs /pr/).
 
     \item \textbf{Archaic Language Failure (curse of the dative}:
     \textit{Ελληνίς} / \textit{ουρανοίς}.
     This includes Katharevousa forms, in specific a nominative in ίς that rhymes with a plural dative in οίς (both morphological forms largely absent in SMG). Surprisingly, Llama 3.3 (Structured) was the only model to perfectly identify this rhyme. The rest struggled with the archaic spelling variances.
 
     \item \textbf{Imperfect Detection}:
     \textit{γρήγορο} / \textit{είσοδο}.
     Imperfect F3 rhyme. No model perfectly captured the full feature set. Claude 4.5 (CoT) and Claude 3.7 (Structured) correctly identified the rhyme type (F3) but missed the specific imperfection details.
 
     \item \textbf{Proper Noun Distraction}:
     \textit{νιότα} / \textit{Ευρώτα}.
     Llama 3.1 70B (Structured) outputted a format error ('STEP'). Claude 4.5, GPT-4o, and Gemini 2.0 generally handled the entity correctly in CoT modes, identifying the rhyme type (F2), besides the challenging synizisis of the first word. 
 
     \item \textbf{Visual vs Phonetic}:
     \textit{ορθός} / \textit{φως}.
     Only  Claude 4.5 (CoT+RAG) got the rhyme correct, a fact that might point to more robust grapheme-to-phoneme mapping capabilities despite the visual mismatch.
 \end{enumerate}
 
 \subsection{Rhyme Generation Results}

 \begin{table}[t]
\centering
\small
\begin{tabular}{lccc}
\toprule
\textbf{Model} & \textbf{No Verify} & \textbf{With Verify} & \textbf{Improvement} \\
\midrule
Claude 4.5 & 0.0\% & 34.6\% & +34.6\% \\
Claude 3.7 & 3.8\% & \textbf{73.1\%} & +69.3\% \\
GPT-4o & 0.0\% & 42.3\% & +42.3\% \\
\bottomrule
\end{tabular}
\caption{Generation validity (\% of perfectly valid poems) with and without phonological verification loop.}
\label{tab:generation}
\end{table}
 
 Table~\ref{tab:generation_by_feature} shows generation validity broken down by feature complexity.
 
 \begin{table}[t]
 \centering
 \small
 \resizebox{\columnwidth}{!}{
 \begin{tabular}{lcccc}
 \toprule
 \textbf{Feature Type} & \textbf{n} & Claude 3.7 & Claude 4.5 & GPT-4o \\
 \midrule
 BASIC (no features) & 6 & \textbf{100.0\%} & 83.3\% & 90.9\% \\
 IMPERFECT & 2 & 50.0\% & 50.0\% & 33.3\% \\
 IDV+PURE & 2 & 50.0\% & 0.0\% & 66.7\% \\
 IDV+RICH & 2 & 50.0\% & 0.0\% & 0.0\% \\
 IDV+IMPERFECT & 2 & \textbf{100.0\%} & 0.0\% & \textbf{100.0\%} \\
 IDV+MOSAIC+PURE & 2 & 0.0\% & 0.0\% & 0.0\% \\
 \bottomrule
 \end{tabular}
 }
 \caption{Generation validity (\%) with verification, by feature complexity. BASIC rhymes (no features) achieve highest success. Complex multi-feature combinations (e.g., IDV+MOSAIC+PURE) fail even with verification.}
 \label{tab:generation_by_feature}
 \end{table}
 
 The verification loop greatly enhances rhyme  validity across all models. Claude 3.7 achieves the highest verified generation rate (73.1\% valid poems), demonstrating that the hybrid approach successfully compensates for LLM phonological weaknesses. Critically, pure LLM generation fails catastrophically (0-4\% validity), confirming that deterministic verification is essential for constrained poetry generation, at least for the models used. 
 
 \subsection{Qualitative Analysis of Generation}
Three main types of generation errors (full output traces are provided in Appendix B) are generally corrected by the loop:
 
 \begin{enumerate}
     \item \textbf{Correcting the stress pattern(F3 vs F2)}:
     When asked for F3 (proparoxytone) rhymes, models are often incorrectly  defaulting to F2 (paroxytone).
     \textit{Example}: GPT-4o initially generated \textit{αναβιώνει} / \textit{απλώνει} (F2). The verifier returned \texttt{"Stress mismatch: Expected F3, found F2"}. The model corrected this to a valid F3 rhyme in the subsequent iteration.
 
     \item \textbf{Feature Precision (Rich vs Pure)}:
     LLMs often treat rhyme types loosely. When M-PURE was requested, Claude 3.7 generated \textit{σκληρό} / \textit{θησαυρό} (which is RICH, sharing the /r/ onset). 
     After the intervention of the loop, the model successfully refined the output to a strict PURE rhyme, demonstrating the system's ability to enforce precise stylistic constraints.
 
     \item \textbf{Non-rhymes}:
     In some cases, Claude 4.5  proposed non-rhyming outputs, e.g. pairs like \textit{αγαπώ} / \textit{ξέρω}. The verification loop noted these invalid pairs (\texttt{"No rhyme"}), and forced the model to regenerate valid phonological matches.
 \end{enumerate}
 
 \section{Discussion}
 
 \subsection{Why LLMs Struggle with Rhyme}
 
 One issue is potentially tied to the nature of LLM tokenizers (e.g., BPE, WordPiece), i.e. that fact that they segment text based on statistical co-occurrence and not linguistically motivated units. In Greek, this often results in words being split mid-syllable or mid-grapheme-cluster, entirely obscuring the phonological structure required for rhyme. For example, a word like \textit{παράπονο} (pa-'ra-po-no) might be tokenized as [πα, ρά, πο, νο] in an ideal case, but often appears as [παρ, άπ, ονο] depending on the vocabulary, stripping the model of the ability to map the stress position (F3) relative to the final syllable.
 
 Unlike orthographical systems where text maps close to 1:1 to sound, Greek's retention of historical orthography, i.e. a system of orthography that has been retained for millennia and, thus, has not followed the changes in the language,  features many-to-one mappings (e.g., the sound /i/ can be spelled as $\iota, \eta, v, \epsilon\iota, o\iota, v\iota$). LLMs trained primarily on text often rely on visual similarity ("eye rhyme") rather than phonetic identity. This explains why models fail on \textit{κρίνοι/κρίνει} (visual mismatch but phonetic match) while hallucinating rhymes for \textit{ορθός/φως} (visual mismatch and phonetic mismatch).
 
 Finally, note that Greek rhyme is strictly defined by stress position (M, F2, F3). Since LLMs lack an explicit prosodic module, they are prone to fail when attempting to distinguish minimal pairs that differ only in stress (e.g., \textit{νόμος} vs \textit{νομός}).
 
 \subsection{Creativity vs. Hallucination}
 
 Our analysis of "hallucinated" words reveals a nuanced trade-off between semantic grounding and poetic creativity. We observe two distinct categories of invented vocabulary.
 
 Some of the hallucinations can be very well-taken as poetic neologism. For example, Claude 4.5 gives us \textit{αιωνημένα} (eternal-ized) and \textit{θάρραμα} (courage-thing), which are not only valid phonotactically, but also are meaningful enough neologisms (for eample αιωνημένα can be an adjective that means something related to enternity, while θάρραμα can be seen as a collage of θάρρος (courage) and χάραμα (dawn).  A striking example is \textit{πυραφί} (fire-colored), which appears to be constructed by analogy to \textit{χρυσαφί} (gold-colored), demonstrating a deep (if unauthorized) grasp of Greek morphology. Similarly, \textit{αγέρη} appears as a valid homophone of \textit{αγέρι} (breeze), suggesting a spelling variation rather than a failure. Nonsense Failures, on the other hand, are phonotactically correct words but  are difficult to coerce into a meaningful word. Examples include \textit{τσίγκλο} and \textit{σκάμπουνε}, which lack semantic transparency. It is quite positive that no phonotactic violations of Greek happen in these cases (modulo the total English nonsense generated by some Llama models). 
 
 We find a clear, we would dare to call it a ``personality'' difference in models: GPT-4o is the "safest" model, producing almost zero nonsense but also fewer neologisms. In contrast, Claude 4.5 is the boldest,  with a high rate of invention, the majority of which are plausible neologisms rather than nonsense. This suggests that what is often penalized as "hallucination" in factual tasks can serve as a proxy for creativity in poetic tasks.
 
 Crucially, this higher level of invention coincides with stronger performance in the verification loop. It appears that the strict constraints imposed by the verifier push capable models to neologize in order to conform to the poetic form, inventing new words when standard vocabulary fails to meet the phonological requirements.
 
 \subsection{Benefits of Hybrid Architecture}
  The hybrid approach succeeds by decoupling reasoning from phonology. The deterministic engine provides the ground truth that the neural model optimizes towards. This "Sandwich" architecture (Prompt $\rightarrow$ LLM $\rightarrow$ Verify $\rightarrow$ Refine) allows us to leverage the semantic richness and creativity of large language models while imposing the strict phonological constraints required by the poetic form. By offloading the "scoring" of a rhyme to a precise symbolic module, we free the LLM to focus on lexical selection and thematic coherence, effectively bypassing its inherent blindness to sub-token phonological structures.
 
  \section{Future Work}
 
  The natural next step for such a system is to go beyond rhyme and implemenent metrical verification. Greek poetry relies heavily on stress-timed constituent meters (e.g., iambic 15-syllable verse) and systems that could handle both rhyme and metrical structure would be a very interesting research direction to take. 
 
  Additionally, we plan to explore Reinforcement Learning from Phonological Feedback (RLPF). Instead of a simple rejection sampling loop at inference time, the signals from our deterministic verifier could be used as a reward function to fine-tune a smaller model (e.g., Llama 8B), potentially internalizing phonological constraints directly into the model's weights.
 
  Finally, we would hope to extend this hybrid neuro-symbolic approach beyond Greek to other low-resource languages, the goal being to create a generalized "Universal Rhyme Engine" that requires only a language-specific phonological rule set.
 
 \section{Conclusion}
 
 In this work, we presented a hybrid neuro-symbolic system designed to bridge the gap between Large Language Models and strict phonological constraints in Modern Greek poetry. Our experiments reveal that while LLMs possess latent creative capabilities, they fundamentally struggle with the precise phonological computations required for rigorous rhyme detection and generation, particularly in lower-resource languages. By integrating a deterministic phonological engine with an agentic generation loop, we demonstrated a dramatic improvement in generation validity, raising success rates from a baseline of under 4\% to 73.1\%. Furthermore, our identification benchmarks exposed a significant "Reasoning Gap," where only the most advanced reasoning models (using Chain-of-Thought) could compete with symbolic verification. We hope that our released codebase and the curated corpus of 40,000+ Modern Greek rhymes will serve as foundational resources for future research into phonologically-aware NLP, suggesting that hybrid architectures remain essential for mastering the structural nuances of poetic form.
 
 \section*{Limitations}
 The size of our filtered corpus, while ensuring high phonological quality, remains significantly smaller than equivalent datasets available for high-resource languages like English. Second, the proposed hybrid architecture imposes a computational overhead; the iterative nature of the Generate-Verify-Refine loop inevitably increases the total generation time compared to standard single-pass LLM inference.

  \section*{Acknowledgments}
  We gratefully acknowledge the Centre for the Greek Language (Κέντρο Ελληνικής Γλώσσας) for providing access to the Anemoskala corpus and granting permission to derive our rhyming dataset from their digital resources. The original corpus is available through the Portal for the Greek Language (www.greek-language.gr).

 \bibliography{custom}
 
 \appendix
 \onecolumn
\section{Appendix: Detailed Qualitative Results}

This appendix presents the full raw outputs for the 9 representative test cases discussed in the Qualitative Analysis. For each case, we show the predicted Rhyme Type and Features for all 8 models across 4 configurations (Structured vs CoT, No RAG vs RAG).

\textbf{Legend}: \textcolor{green}{\checkmark} = Correct Type \& Features, \textcolor{orange}{\warning} = Correct Type but Wrong Features, \textcolor{red}{\xmark} = Incorrect Type.

\subsection{1. Baseline Success (M) (Keyword: 'απαιτώ')}
\textbf{Poem}: \textit{Έτσι από σένα περιμένω κι απαιτώ. / της Τραγωδίας τον Λόγο τον λαμπρό —}\\
\textbf{True Label}: M ['PURE']

\begin{table}[H]
\centering
\scriptsize
\begin{tabular}{l|c|c|c|c}
\toprule
Model & Struct (No RAG) & Struct (RAG) & CoT (No RAG) & CoT (RAG) \\
\midrule
Claude 4.5 & \textcolor{orange}{\warning} M & \textcolor{green}{\checkmark} M {[}PURE{]} & \textcolor{red}{\xmark} S & \textcolor{orange}{\warning} M \\
Claude 3.7 & \textcolor{red}{\xmark} MISS & \textcolor{red}{\xmark} MISS & \textcolor{orange}{\warning} M & \textcolor{red}{\xmark} I \\
GPT-4o & \textcolor{red}{\xmark} MISS & \textcolor{orange}{\warning} M {[}IMPERFECT{]} & \textcolor{orange}{\warning} M & \textcolor{red}{\xmark} S \\
Gemini 2.0 & \textcolor{orange}{\warning} M {[}IMPERFECT{]} & \textcolor{orange}{\warning} M {[}IMPERFECT{]} & \textcolor{red}{\xmark} MISS & \textcolor{red}{\xmark} MISS \\
Llama 70B & \textcolor{orange}{\warning} M {[}IDV, RICH{]} & \textcolor{red}{\xmark} F2 {[}PURE{]} & \textcolor{red}{\xmark} F2 & \textcolor{red}{\xmark} + \\
Llama 3.3 & \textcolor{red}{\xmark} S & \textcolor{orange}{\warning} M {[}IMPERFECT{]} & \textcolor{red}{\xmark} MISS & \textcolor{red}{\xmark} OF \\
Mistral & \textcolor{orange}{\warning} M {[}IMPERFECT{]} & \textcolor{red}{\xmark} STRESS & \textcolor{red}{\xmark} S & \textcolor{red}{\xmark} IN \\
Llama 8B & \textcolor{red}{\xmark} S & \textcolor{red}{\xmark} MISS & \textcolor{red}{\xmark} WORKERJOE & \textcolor{red}{\xmark} ANN \\
\bottomrule
\end{tabular}
\end{table}

\subsection{2. Baseline Success (F2) (Keyword: 'κρίνοι')}
\textbf{Poem}: \textit{Από ρουμπίνια ρόδα, από μαργαριτάρια κρίνοι, / από αμεθύστους μενεξέδες. Ως αυτός τα κρίνει,}\\
\textbf{True Label}: F2 ['RICH']

\begin{table}[H]
\centering
\scriptsize
\begin{tabular}{l|c|c|c|c}
\toprule
Model & Struct (No RAG) & Struct (RAG) & CoT (No RAG) & CoT (RAG) \\
\midrule
Claude 4.5 & \textcolor{orange}{\warning} F2 {[}COPY, RICH{]} & \textcolor{red}{\xmark} BETWEEN & \textcolor{red}{\xmark} S & \textcolor{orange}{\warning} F2 \\
Claude 3.7 & \textcolor{orange}{\warning} F2 {[}COPY{]} & \textcolor{orange}{\warning} F2 {[}COPY{]} & \textcolor{red}{\xmark} M & \textcolor{red}{\xmark} TR {[}CC{]} \\
GPT-4o & \textcolor{red}{\xmark} S & \textcolor{red}{\xmark} S & \textcolor{orange}{\warning} F2 & \textcolor{red}{\xmark} MASCULINE \\
Gemini 2.0 & \textcolor{red}{\xmark} MISS & \textcolor{red}{\xmark} MISS & \textcolor{red}{\xmark} S & \textcolor{red}{\xmark} MISS \\
Llama 70B & \textcolor{orange}{\warning} F2 {[}2W, C2, IDV, RICH{]} & \textcolor{green}{\checkmark} F2 {[}RICH{]} & \textcolor{red}{\xmark} S & \textcolor{red}{\xmark} TR \\
Llama 3.3 & \textcolor{red}{\xmark} MISS & \textcolor{orange}{\warning} F2 {[}PURE{]} & \textcolor{red}{\xmark} M & \textcolor{red}{\xmark} MISS \\
Mistral & \textcolor{red}{\xmark} STRESS & \textcolor{orange}{\warning} F2 {[}CC, RICH{]} & \textcolor{red}{\xmark} S & \textcolor{red}{\xmark} 2 \\
Llama 8B & \textcolor{red}{\xmark} S & \textcolor{red}{\xmark} RECEIVE & \textcolor{red}{\xmark} CLIENT & \textcolor{red}{\xmark} M \\
\bottomrule
\end{tabular}
\end{table}

\subsection{3. Structural Failure (F3) (Keyword: 'παράπονο')}
\textbf{Poem}: \textit{αρχίζει το μωρό ένα παράπονο, / που ήρθε σ’ έναν κόσμο τόσον άπονο.}\\
\textbf{True Label}: F3 ['PURE']

\begin{table}[H]
\centering
\scriptsize
\begin{tabular}{l|c|c|c|c}
\toprule
Model & Struct (No RAG) & Struct (RAG) & CoT (No RAG) & CoT (RAG) \\
\midrule
Claude 4.5 & \textcolor{red}{\xmark} MISS & \textcolor{red}{\xmark} IN & \textcolor{red}{\xmark} S & \textcolor{red}{\xmark} NOT \\
Claude 3.7 & \textcolor{red}{\xmark} MISS & \textcolor{orange}{\warning} F3 {[}MOSAIC{]} & \textcolor{orange}{\warning} F3 & \textcolor{orange}{\warning} F3 {[}IMPERFECT{]} \\
GPT-4o & \textcolor{red}{\xmark} MISS & \textcolor{red}{\xmark} MISS & \textcolor{red}{\xmark} MATCH & \textcolor{orange}{\warning} F3 {[}IMPERFECT, RICH{]} \\
Gemini 2.0 & \textcolor{red}{\xmark} MISS & \textcolor{orange}{\warning} F3 {[}IDV, RICH{]} & \textcolor{red}{\xmark} MISS & \textcolor{red}{\xmark} F2 \\
Llama 70B & \textcolor{red}{\xmark} MISS & \textcolor{red}{\xmark} F2 {[}PURE{]} & \textcolor{red}{\xmark} F2 & \textcolor{red}{\xmark} M \\
Llama 3.3 & \textcolor{red}{\xmark} M {[}C1, RICH{]} & \textcolor{red}{\xmark} MISS & \textcolor{red}{\xmark} S & \textcolor{red}{\xmark} MISS \\
Mistral & \textcolor{red}{\xmark} STRESS & \textcolor{red}{\xmark} F2 {[}IDV, PURE{]} & \textcolor{red}{\xmark} S & \textcolor{red}{\xmark} MISS \\
Llama 8B & \textcolor{red}{\xmark} K & \textcolor{red}{\xmark} F2 {[}IDV, RICH{]} & \textcolor{red}{\xmark} S & \textcolor{red}{\xmark} ZIGSPACE \\
\bottomrule
\end{tabular}
\end{table}

\subsection{4. Mosaic Failure (Keyword: 'λυγμέ')}
\textbf{Poem}: \textit{του αθηναίικου εσύ χινόπωρου λυγμέ, / ψιχάλα κυνηγάρα, που έβρεχες για με;}\\
\textbf{True Label}: M ['MOSAIC']

\begin{table}[H]
\centering
\scriptsize
\begin{tabular}{l|c|c|c|c}
\toprule
Model & Struct (No RAG) & Struct (RAG) & CoT (No RAG) & CoT (RAG) \\
\midrule
Claude 4.5 & \textcolor{orange}{\warning} M {[}COPY{]} & \textcolor{orange}{\warning} M {[}COPY{]} & \textcolor{orange}{\warning} M {[}COPY{]} & \textcolor{orange}{\warning} M {[}MOSAIC, PURE{]} \\
Claude 3.7 & \textcolor{orange}{\warning} M {[}IMPERFECT{]} & \textcolor{green}{\checkmark} M {[}MOSAIC{]} & \textcolor{red}{\xmark} S & \textcolor{orange}{\warning} M {[}PURE{]} \\
GPT-4o & \textcolor{red}{\xmark} S & \textcolor{red}{\xmark} MISS & \textcolor{orange}{\warning} M {[}COPY{]} & \textcolor{orange}{\warning} M \\
Gemini 2.0 & \textcolor{red}{\xmark} MISS & \textcolor{orange}{\warning} M {[}IMPERFECT{]} & \textcolor{red}{\xmark} MISS & \textcolor{orange}{\warning} M \\
Llama 70B & \textcolor{red}{\xmark} S & \textcolor{red}{\xmark} MISS & \textcolor{red}{\xmark} S & \textcolor{red}{\xmark} NO \\
Llama 3.3 & \textcolor{red}{\xmark} GIVEN & \textcolor{orange}{\warning} M {[}IMPERFECT{]} & \textcolor{red}{\xmark} S & \textcolor{red}{\xmark} MISS \\
Mistral & \textcolor{red}{\xmark} STRESS & \textcolor{red}{\xmark} STRESS & \textcolor{orange}{\warning} M {[}IMPERFECT, RICH{]} & \textcolor{orange}{\warning} M {[}IMPERFECT{]} \\
Llama 8B & \textcolor{orange}{\warning} M & \textcolor{red}{\xmark} MISS & \textcolor{red}{\xmark} SATURDAY & \textcolor{red}{\xmark} S \\
\bottomrule
\end{tabular}
\end{table}

\subsection{5. Feature Hallucination (Keyword: 'αφρός')}
\textbf{Poem}: \textit{Γιασεμιά, και κοράκια. Και των άσπρων ο αφρός / και του μαύρου η φοβέρα πάντα εντός μου κι εμπρός.}\\
\textbf{True Label}: M ['PURE']

\begin{table}[H]
\centering
\scriptsize
\begin{tabular}{l|c|c|c|c}
\toprule
Model & Struct (No RAG) & Struct (RAG) & CoT (No RAG) & CoT (RAG) \\
\midrule
Claude 4.5 & \textcolor{orange}{\warning} M {[}IMPERFECT{]} & \textcolor{orange}{\warning} M {[}C1, RICH{]} & \textcolor{orange}{\warning} M & \textcolor{green}{\checkmark} M {[}PURE{]} \\
Claude 3.7 & \textcolor{red}{\xmark} S & \textcolor{orange}{\warning} M {[}RICH{]} & \textcolor{orange}{\warning} M & \textcolor{red}{\xmark} IN \\
GPT-4o & \textcolor{red}{\xmark} S & \textcolor{red}{\xmark} IN & \textcolor{orange}{\warning} M & \textcolor{red}{\xmark} IN \\
Gemini 2.0 & \textcolor{orange}{\warning} M {[}C2, RICH{]} & \textcolor{green}{\checkmark} M {[}PURE{]} & \textcolor{orange}{\warning} M & \textcolor{red}{\xmark} S \\
Llama 70B & \textcolor{orange}{\warning} M {[}IMPERFECT{]} & \textcolor{red}{\xmark} MISS & \textcolor{red}{\xmark}  {[}IA{]} & \textcolor{red}{\xmark} F2 \\
Llama 3.3 & \textcolor{red}{\xmark} S & \textcolor{red}{\xmark} OF & \textcolor{red}{\xmark} IN & \textcolor{orange}{\warning} M \\
Mistral & \textcolor{red}{\xmark} STRESS & \textcolor{red}{\xmark} MISS & \textcolor{orange}{\warning} M {[}F2, IMPERFECT{]} & \textcolor{orange}{\warning} M {[}IMPERFECT{]} \\
Llama 8B & \textcolor{red}{\xmark} MISS & \textcolor{red}{\xmark} F2 {[}MOSAIC{]} & \textcolor{red}{\xmark} STRING & \textcolor{red}{\xmark} COMPLETE \\
\bottomrule
\end{tabular}
\end{table}

\subsection{6. Archaic Failure (Keyword: 'Ελληνίς')}
\textbf{Poem}: \textit{Την εγέννησεν είς δήμος, μία πόλις Ελληνίς, / αλλ’ ευθύς εκείνη έπτη, κι έστησεν εν ουρανοίς}\\
\textbf{True Label}: M ['RICH']

\begin{table}[H]
\centering
\scriptsize
\begin{tabular}{l|c|c|c|c}
\toprule
Model & Struct (No RAG) & Struct (RAG) & CoT (No RAG) & CoT (RAG) \\
\midrule
Claude 4.5 & \textcolor{red}{\xmark} MISS & \textcolor{red}{\xmark} S & \textcolor{orange}{\warning} M & \textcolor{orange}{\warning} M {[}PURE{]} \\
Claude 3.7 & \textcolor{red}{\xmark} F2 {[}RICH{]} & \textcolor{orange}{\warning} M {[}PURE{]} & \textcolor{red}{\xmark} S & \textcolor{orange}{\warning} M \\
GPT-4o & \textcolor{red}{\xmark} OF & \textcolor{red}{\xmark} OF & \textcolor{orange}{\warning} M {[}IDENTICAL{]} & \textcolor{red}{\xmark} S \\
Gemini 2.0 & \textcolor{orange}{\warning} M {[}IMPERFECT{]} & \textcolor{red}{\xmark} MISS & \textcolor{orange}{\warning} M & \textcolor{red}{\xmark} MISS \\
Llama 70B & \textcolor{red}{\xmark} ING & \textcolor{red}{\xmark} MISS & \textcolor{orange}{\warning} M & \textcolor{red}{\xmark} NO \\
Llama 3.3 & \textcolor{green}{\checkmark} M {[}RICH{]} & \textcolor{red}{\xmark} F2 {[}PURE{]} & \textcolor{red}{\xmark} BOTH & \textcolor{red}{\xmark} OF \\
Mistral & \textcolor{red}{\xmark} MISS & \textcolor{red}{\xmark} F2 {[}C1, IMPERFECT, RICH{]} & \textcolor{red}{\xmark} S & \textcolor{red}{\xmark} MISS \\
Llama 8B & \textcolor{red}{\xmark} IRSECURITY & \textcolor{red}{\xmark} DAV & \textcolor{red}{\xmark} CHOOSING & \textcolor{red}{\xmark} MISS \\
\bottomrule
\end{tabular}
\end{table}

\subsection{7. Imperfect Detection (Keyword: 'γρήγορο')}
\textbf{Poem}: \textit{αόριστη, με διάβα γρήγορο, / Στου καφενείου την είσοδο}\\
\textbf{True Label}: F3 ['IMP', 'C', 'IMPERFECT']

\begin{table}[H]
\centering
\scriptsize
\begin{tabular}{l|c|c|c|c}
\toprule
Model & Struct (No RAG) & Struct (RAG) & CoT (No RAG) & CoT (RAG) \\
\midrule
Claude 4.5 & \textcolor{red}{\xmark} S & \textcolor{red}{\xmark} S & \textcolor{orange}{\warning} F3 {[}IMPERFECT{]} & \textcolor{red}{\xmark} IMPERFECT \\
Claude 3.7 & \textcolor{orange}{\warning} F3 {[}C1, IMPERFECT, RICH{]} & \textcolor{red}{\xmark} S & \textcolor{red}{\xmark} F2 & \textcolor{red}{\xmark} I \\
GPT-4o & \textcolor{orange}{\warning} F3 {[}IMPERFECT{]} & \textcolor{red}{\xmark} F2 {[}IMPERFECT{]} & \textcolor{red}{\xmark} F2 {[}F2{]} & \textcolor{red}{\xmark} S \\
Gemini 2.0 & \textcolor{red}{\xmark} MISS & \textcolor{orange}{\warning} F3 {[}IMPERFECT{]} & \textcolor{red}{\xmark} MISS & \textcolor{red}{\xmark} F2 \\
Llama 70B & \textcolor{red}{\xmark} MISS & \textcolor{red}{\xmark} F2 {[}IMPERFECT{]} & \textcolor{red}{\xmark} M {[}IMPERFECT{]} & \textcolor{red}{\xmark} S \\
Llama 3.3 & \textcolor{red}{\xmark} MISS & \textcolor{red}{\xmark} M {[}IMPERFECT{]} & \textcolor{red}{\xmark} MISS & \textcolor{red}{\xmark} MISS \\
Mistral & \textcolor{red}{\xmark} STRESS & \textcolor{red}{\xmark} STRESS & \textcolor{red}{\xmark} F2 {[}IMPERFECT{]} & \textcolor{red}{\xmark} STRESS \\
Llama 8B & \textcolor{red}{\xmark} F2 {[}2W, IDV{]} & \textcolor{red}{\xmark} MISS & \textcolor{red}{\xmark} D & \textcolor{red}{\xmark} S \\
\bottomrule
\end{tabular}
\end{table}

\subsection{8. Proper Noun Failure (Keyword: 'νιότα')}
\textbf{Poem}: \textit{Πολεμιστή, τα γαληνά σου νιότα / Όπου έπεσες, κλωνάρια κι απ’ του Ευρώτα}\\
\textbf{True Label}: F2 ['PURE']

\begin{table}[H]
\centering
\scriptsize
\begin{tabular}{l|c|c|c|c}
\toprule
Model & Struct (No RAG) & Struct (RAG) & CoT (No RAG) & CoT (RAG) \\
\midrule
Claude 4.5 & \textcolor{orange}{\warning} F2 & \textcolor{red}{\xmark} STEP & \textcolor{orange}{\warning} F2 & \textcolor{orange}{\warning} F2 \\
Claude 3.7 & \textcolor{orange}{\warning} F2 {[}IMPERFECT{]} & \textcolor{red}{\xmark} S & \textcolor{red}{\xmark} F3 & \textcolor{red}{\xmark} I \\
GPT-4o & \textcolor{red}{\xmark} S & \textcolor{red}{\xmark} IN & \textcolor{orange}{\warning} F2 & \textcolor{red}{\xmark} M {[}IDV, PURE{]} \\
Gemini 2.0 & \textcolor{red}{\xmark} MISS & \textcolor{red}{\xmark} MISS & \textcolor{orange}{\warning} F2 & \textcolor{orange}{\warning} F2 \\
Llama 70B & \textcolor{red}{\xmark} MISS & \textcolor{red}{\xmark} MISS & \textcolor{orange}{\warning} F2 & \textcolor{orange}{\warning} F2 \\
Llama 3.3 & \textcolor{red}{\xmark} MISS & \textcolor{red}{\xmark} THE & \textcolor{red}{\xmark} SINCE & \textcolor{red}{\xmark} BUT \\
Mistral & \textcolor{red}{\xmark} STRESS & \textcolor{red}{\xmark} MISS & \textcolor{red}{\xmark} S & \textcolor{red}{\xmark} 3 \\
Llama 8B & \textcolor{red}{\xmark} S & \textcolor{red}{\xmark} F3 {[}2W, CC, IDV, RICH{]} & \textcolor{red}{\xmark} THE & \textcolor{red}{\xmark} F3 \\
\bottomrule
\end{tabular}
\end{table}

\subsection{9. Visual vs Phonetic (Keyword: 'ορθός')}
\textbf{Poem}: \textit{άξαφνα το παράθυρο και στάθηκα ορθός, / τις μυρωδιές, τα χρώματα και το ιλαρό το φως.}\\
\textbf{True Label}: M ['MOSAIC', 'IDV']

\begin{table}[H]
\centering
\scriptsize
\begin{tabular}{l|c|c|c|c}
\toprule
Model & Struct (No RAG) & Struct (RAG) & CoT (No RAG) & CoT (RAG) \\
\midrule
Claude 4.5 & \textcolor{orange}{\warning} M {[}COPY{]} & \textcolor{orange}{\warning} M {[}IMPERFECT{]} & \textcolor{orange}{\warning} M {[}IDV, IMPERFECT{]} & \textcolor{orange}{\warning} M {[}PURE{]} \\
Claude 3.7 & \textcolor{red}{\xmark} MISS & \textcolor{orange}{\warning} M {[}IMPERFECT{]} & \textcolor{red}{\xmark} I & \textcolor{red}{\xmark} IN \\
GPT-4o & \textcolor{red}{\xmark} S & \textcolor{red}{\xmark} S & \textcolor{red}{\xmark} MISS & \textcolor{orange}{\warning} M {[}PURE{]} \\
Gemini 2.0 & \textcolor{red}{\xmark} MISS & \textcolor{orange}{\warning} M {[}IMPERFECT{]} & \textcolor{red}{\xmark} MISS & \textcolor{red}{\xmark} MISS \\
Llama 70B & \textcolor{red}{\xmark} S & \textcolor{red}{\xmark} MISS & \textcolor{red}{\xmark} S & \textcolor{orange}{\warning} M {[}IMPERFECT{]} \\
Llama 3.3 & \textcolor{orange}{\warning} M {[}IMPERFECT{]} & \textcolor{orange}{\warning} M {[}IMPERFECT{]} & \textcolor{red}{\xmark} MISS & \textcolor{red}{\xmark} IN \\
Mistral & \textcolor{red}{\xmark} MISS & \textcolor{orange}{\warning} M {[}PURE{]} & \textcolor{red}{\xmark} S & \textcolor{red}{\xmark} ONSET \\
Llama 8B & \textcolor{red}{\xmark} MISS & \textcolor{red}{\xmark} THE & \textcolor{red}{\xmark} BETWEEN & \textcolor{red}{\xmark} S \\
\bottomrule
\end{tabular}
\end{table}
\clearpage
\section{Appendix: Detailed Generation Traces}
We present full interaction traces for the three classes of generation errors discussed in the Qualitative Analysis. These examples demonstrate how the hybrid verification loop intercepts compliant but incorrect outputs (Stress, Feature Precision) and hallucinations, guiding the model to phonologically valid results.

\subsection{1. Stress Pattern Correction (GPT-4o)}
\textbf{Test Case}: Theme "Sky" (\textit{ουρανός}), Constraint: \textbf{F3 (Proparoxytone) Rhyme}.\\
The model initially defaults to the more common F2 pattern.

\noindent\textbf{Initial Draft (Pure LLM)}:
\begin{quote}
\textit{Στο χέρι σου το μυστικό αναβιώνει,}\\ 
\textit{και μέσα στα σκιερά πελάγη απλώνει.}
\end{quote}
\textbf{Verifier Feedback}: \texttt{Stress mismatch: Expected F3, found F2 for 'αναβιώνει'/'απλώνει'.}

\noindent\textbf{Refined Output (Hybrid)}:
\begin{quote}
\textit{Ω θεέ μου, στον ουρανό να ψιθυρίσουμε,}\\
\textit{τα όνειρά μας να πετάξουν να αγγίξουμε.}  
\end{quote}
\textit{Analysis}: The corrected rhymes (\textit{ψιθυρίσουμε} / \textit{αγγίξουμε}) correctly carry stress on the antepenultimate syllable.

\subsection{2. Feature Precision (Claude 3.7)}
\textbf{Test Case}: Theme "Light" (\textit{φως}), Constraint: \textbf{M-PURE Rhyme}.\\
The model generates a valid rhyme that violates the "Pure" constraint (by being "Rich").

\noindent\textbf{Initial Draft (Pure LLM)}:
\begin{quote}
\textit{Θα λάμπει πάντα φωτεινό στον κόσμο το σκληρό,}\\
\textit{γλυκαίνει κάθε σκοτεινή γωνιά σαν θησαυρό.}
\end{quote}
\textbf{Verifier Feedback}: \texttt{PURE requested but 'σκληρό'/'θησαυρό' is RICH (Onset: /r/).}

\noindent\textbf{Refined Output (Hybrid)}:
\begin{quote}
\textit{Το φως σαν κύμα της χαράς ξυπνά}\\
\textit{τα σκοτεινά μου όνειρα φωτίζει καθαρά}
\end{quote}
\textit{Analysis}: The refined pair (\textit{ξυπνά} / \textit{καθαρά}) has distinct onsets (/p/ vs /th/), satisfying the PURE constraint.

\subsection{3. Hallucination Correction (Claude 4.5)}
\textbf{Test Case}: Theme "Love" (\textit{αγάπη}), Constraint: \textbf{M Rhyme}.\\
The model hallucinates rhymes with no phonetic overlap.

\noindent\textbf{Initial Draft (Pure LLM)}:
\begin{quote}
\textit{Η καρδιά μου χτυπά δυνατά για σένα που αγαπώ,}\\
\textit{κάθε στιγμή που περνά χωρίς εσένα είναι πικρή, το ξέρω.}
\end{quote}
\textbf{Verifier Feedback}: \texttt{No rhyme: 'αγαπώ' / 'ξέρω'.}

\noindent\textbf{Refined Output (Hybrid)}:
\begin{quote}
\textit{Μέσα στην καρδιά μου κρύβεται αγαπώ,}\\
\textit{ένα όνειρο γλυκό που με κρατώ,}
\end{quote}
\textit{Analysis}: The model successfully regenerates a valid rhyme pair (\textit{αγαπώ} / \textit{κρατώ}).

 \section{Prompt Templates}
 See source code for full templates.
 
 \end{document}